\pdfoutput=1
\documentclass[11pt,a4paper]{article}
\usepackage[hyperref]{acl2019}
\usepackage{times}
\usepackage{latexsym}
\usepackage{amsmath}
\usepackage{amsfonts}
\usepackage{amssymb}
\usepackage{latexsym}
\usepackage{booktabs}
\usepackage{color} 
\usepackage{setspace}

\usepackage{multirow}

\usepackage{url}
\usepackage{graphicx} 
\usepackage{subfigure}

\usepackage{algorithm}
\usepackage{algorithmic}

\aclfinalcopy 
\def\aclpaperid{113} 

\title{Head-Driven Phrase Structure Grammar Parsing on Penn Treebank}

\author{Junru Zhou \and Hai Zhao\thanks{$\ $  Corresponding author. This paper was partially supported by National
Key Research and Development Program of China (No.
2017YFB0304100) and key projects of Natural Science Foundation of
China (No. U1836222 and No. 61733011)} \\
  Department of Computer Science and Engineering \\
  Key Lab of Shanghai Education Commission \\
  for Intelligent Interaction and Cognitive Engineering \\
  Shanghai Jiao Tong University, Shanghai, China \\
  {\tt zhoujunru@sjtu.edu.cn, zhaohai@cs.sjtu.edu.cn} 
  }
\date{}

\begin{document}
\maketitle
\begin{abstract}

  Head-driven phrase structure grammar
(HPSG) enjoys a uniform formalism representing rich contextual syntactic and even semantic meanings.
This paper makes the first attempt to formulate a simplified HPSG by integrating constituent and
dependency formal representations into head-driven phrase structure.
Then two parsing algorithms are respectively proposed for two converted tree representations, division span and joint span. 
As HPSG encodes both constituent and dependency structure information, the proposed HPSG parsers may be regarded as a sort of joint decoder for both types of structures and thus are evaluated in terms of extracted or converted constituent and dependency parsing trees.
Our parser achieves new state-of-the-art performance for both parsing tasks on Penn Treebank  (PTB) and Chinese Penn Treebank, verifying the effectiveness of joint learning constituent and dependency structures. In details,  we report 96.33 F1 of constituent parsing and 97.20\%  UAS of dependency parsing on PTB.
\end{abstract}

\section{Introduction}

Head-driven phrase structure grammar (HPSG) is a highly lexicalized, constraint-based grammar developed by \cite{pollard1994head}. As opposed to dependency grammar, HPSG is the immediate successor of generalized phrase structure grammar.

HPSG divides language symbols into categories of different types, such as vocabulary, phrases, etc. Each category has different grammar letter information. The complete language symbol which is a complex type feature structure represented by attribute value matrices (AVMs) includes phonological, syntactic, and semantic properties, the valence of the word and interrelationship between various components of the phrase structure.

Meanwhile, the constituent structure of HPSG follows the HEAD FEATURE PRINCIPLE (HFP) \cite{pollard1994head}: \textit{``}the head value of any headed phrase is structure-shared with the HEAD value of the head daughter. The effect of the HFP is to guarantee that headed phrases really are \textit{projections} of their head daughter\textit{"} (p. 34).

\begin{figure}[t!]
    \centering
    \subfigure[Constituent]{
        \label{Fig.sub.1}
        \includegraphics[width=1.2in]{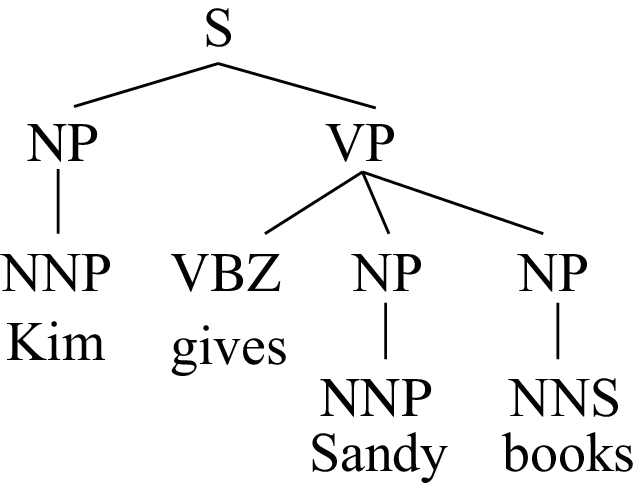}
    }
    \subfigure[Dependency]{
        \label{Fig.sub.2}
        \includegraphics[width=1.6in]{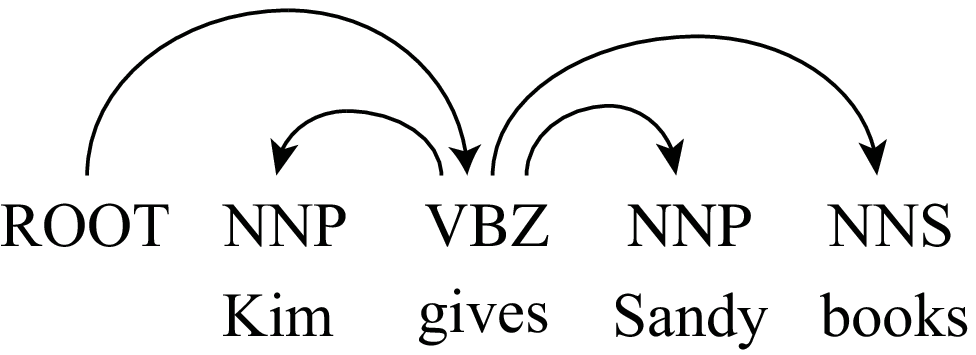}
    }
    \subfigure[HPSG]{
        \label{Fig.sub.3}
        \includegraphics[width=2.5in]{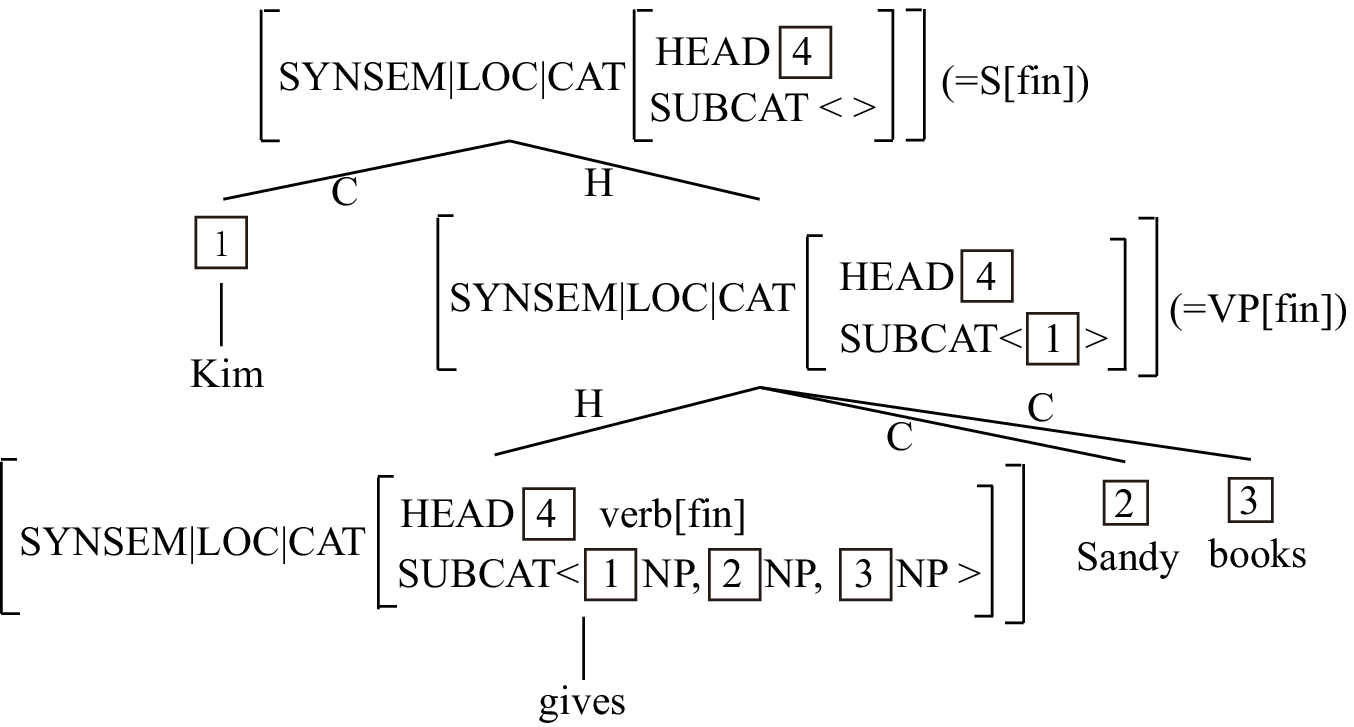}
    }
    \caption{Constituent, dependency and HPSG trees.}
    \label{fig1}
\end{figure}

Constituent and dependency are two typical syntactic structure representation forms, which have been well studied from both linguistic and computational perspective \cite{chomsky1981lectures, Leeman2001Bresnan}. 
The two formalisms carrying distinguished information have each own strengths that constituent structure is better at disclosing phrasal continuity while the dependency structure is better at indicating dependency relation among words.

Typical dependency treebanks are usually converted from constituent treebanks, though they may be independently annotated as well for the same languages. In reverse, constituent parsing can be accurately converted to dependencies representation by grammatical rules or machine learning methods \cite{Marieffe06generatingtyped,MaW10-4146}. 
Such convertibility shows a close relation between constituent and dependency representations, which also have a strong correlation with the HFP of HPSG as shown in Figure \ref{fig1}. 
Thus, it is possible to combine the two representation forms into a simplified HPSG not only for even better parsing but also for more linguistically rich representation.

In this work, we exploit both strengths of the two representation forms and combine them into HPSG. To our best knowledge, it is first attempt to perform such a formulization\footnote{Code and trained English models are publicly available: https://github.com/DoodleJZ/HPSG-Neural-Parser}.
In this paper, we explore two parsing methods for the simplified HPSG parse tree which contains both constituent and dependency syntactic information. 

Our simplified HPSG will be from the annotations or conversions of Penn Treebank (PTB)\footnote{PTB is an English treebank, our parser will also be evaluated on Chinese Penn Treebank (CTB) which follows the similar annotation guideline as PTB.} \cite{MarcusJ93-2004}. Thus the evaluation for our HPSG parser will also be done on both the annotated constituent and converted dependency parse trees, which let our HPSG parser compare to existing constituent and dependency parsers individually.

Our experimental results show that our HPSG parser brings better prediction on both constituent and dependency tree structures.
In addition, the empirical results show that our parser reaches new state-of-the-art for both parsing tasks.
To sum up, we make the following contributions:

\textbullet \ For the first time, we formulate a simplified HPSG by combining constituent and dependency tree structures.

\textbullet \ We propose two novel methods to handle the simplified HPSG parsing.

\textbullet \ Our model achieves state-of-the-art results on PTB and CTB for both constituent and dependency parsing.

The rest of the paper is organized as follows: Section 2 presents the tree structure of HPSG and two span representations. Section 3 presents our model based on self-attention architecture and the adopted parsing algorithms. 
Section 4 reports the experiments and results on PTB and CTB treebanks to evaluate our model. 
At last, we survey related work and conclude this paper respectively in Sections 5 and 6.

\begin{figure}[t!]
    \centering
    \includegraphics[width=2.5in]{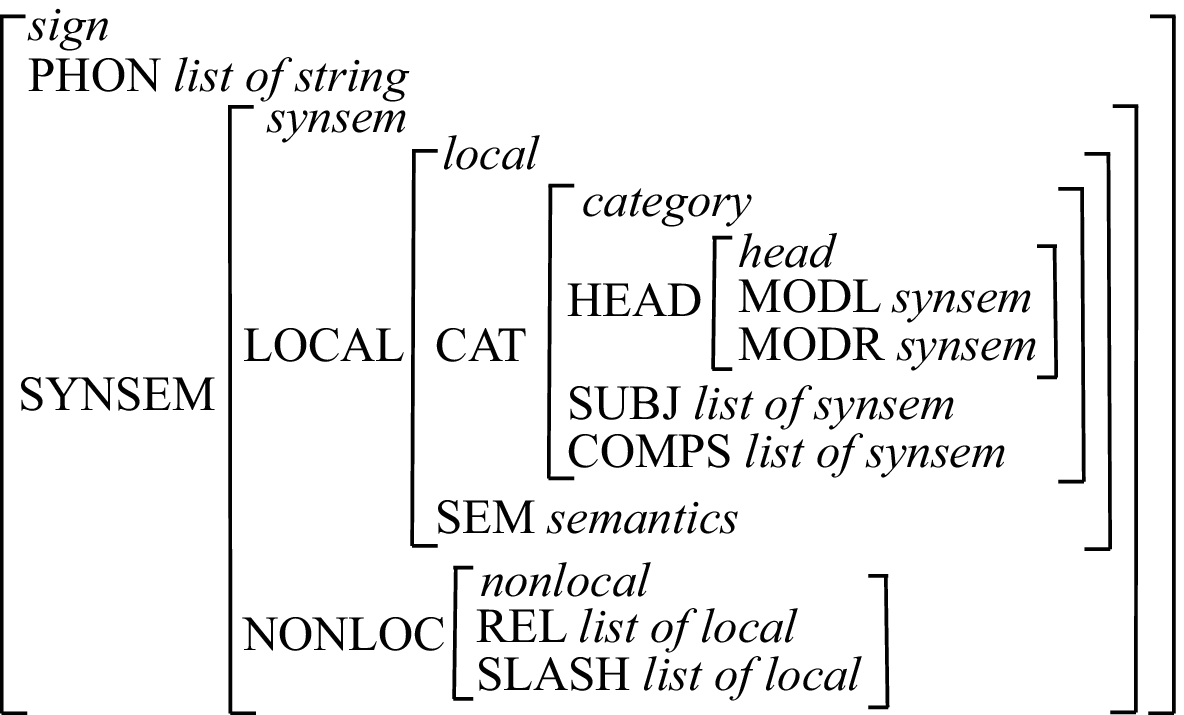}
    
    \caption{HPSG sign from \cite{Miyao2004CorpusOrientedGD}.}
    \label{sign}
\end{figure}

\section{Simplified HPSG on PTB}

\cite{Miyao2004CorpusOrientedGD} reports the first work of semi-automatically acquiring an English HPSG grammar from the Penn Treebank. Figure \ref{sign} demonstrates an HPSG unit presentation (formally called sign), in which head consists of the essential information.    
As the work of \cite{Miyao2004CorpusOrientedGD} cannot demonstrate an accurate enough HPSG from the entire source constituent treebank, we focus on the core of HPSG sign, HEAD, which is conveniently connected with dependency grammar. For the purpose of accurate HPSG building, in this work, we construct a simplified HPSG only from annotations of PTB by combining constituent and dependency parse trees.

\begin{figure*}[t!]
    \centering
    \subfigure[Constituent and dependency.]{
        \label{Fig.sub.1}
        \includegraphics[width=5in]{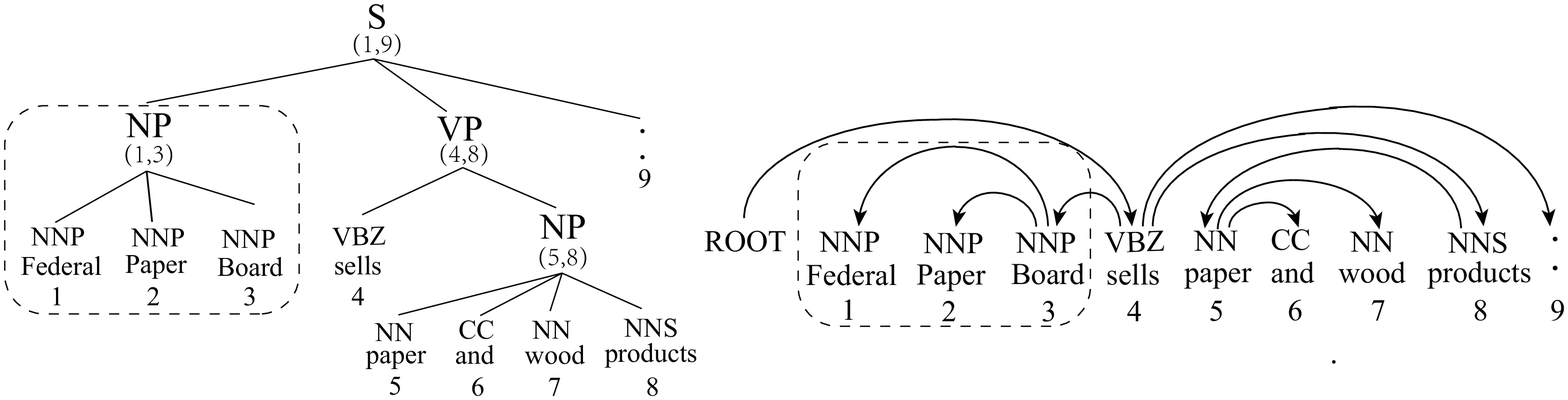}
    }
    \subfigure[Division span structure.]{
        \label{Fig.sub.2}
        \includegraphics[width=2.3in]{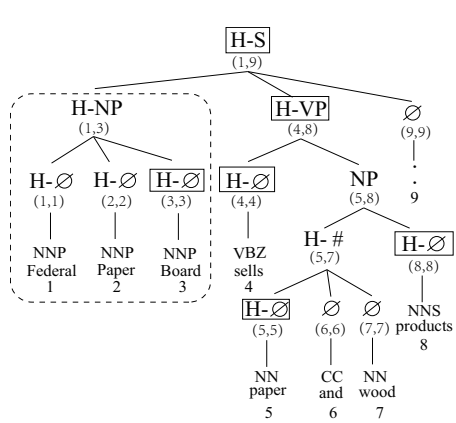}
    }
    \subfigure[Joint span structure.]{
        \label{Fig.sub.3}
        \includegraphics[width=2.3in]{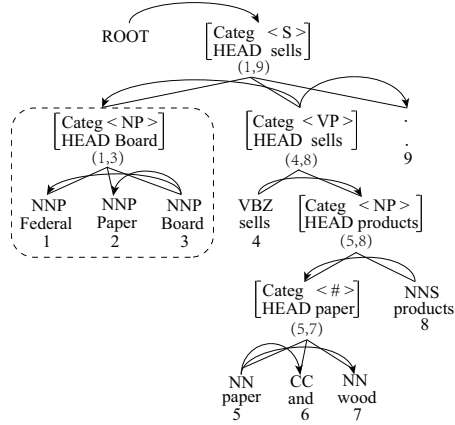}
    }
    \caption{Constituent, dependency and two different simplified HPSG structures of the same sentence which is indexed from 1 to 9 and assigned interval range for each node. Dotted box represents the same part. The special category $\#$ is assigned to divide the phrase with multiple heads.
    Division span structure adds token $H$ in front of the category to distinguish whether the phrase is on the left or right of the head. Thus the head is the last one of the category with $H$ which is marked with a box. Joint span structure contains constitute phrase and dependency arc. \textit{Categ} in each node represents the category of each constituent and \textit{HEAD} indicates the head word.}
    \label{fig2}
\end{figure*}

\subsection{Tree Preprocessing}

In standard HPSG relating to HFP, the HEAD value of any headed phrase is structure-shared with the HEAD value of the head daughter. 
In other words, the phrase in our simplified HPSG tree may be exactly the same as that in a constituent tree and the head word of the phrase corresponding to the parent of the head word of its children in dependency tree\footnote{In standard HPSG, the HEAD value is the part-of-speech of the head word. But in our simplified HPSG tree, we set the head word as HEAD value for convenience.}.
For example, in the constituent tree of Figure \ref{Fig.sub.1}, \textit{Federal Paper Board} is a phrase $(1,3)$ assigned with category NP and in dependency tree, \textit{Board} is parent of \textit{Federal} and \textit{Paper}, thus in our simplified HPSG tree, the head of phrase $(1,3)$ is \textit{Board}. 

For dependency parsing on PTB, the dependency structures are mainly obtained by converting constituent structure with three head rules:
(1) Penn2Malt{\footnote{http://cl.lingfil.uu.se/~nivre/research/Penn2Malt.html}}
and the head rules of Yamada and Matsumoto (2003), noted as PTB-YM;
(2) LTH Converter{\footnote{http://nlp.cs.lth.se/software/treebank\_converter}} \cite{JohanssonW07-2416}, noted as PTB-LTH;
(3) Stanford parser{\footnote{http://nlp.stanford.edu/software/lex-parser.html}}\cite{Marieffe06generatingtyped},  noted as PTB-SD.

Following most of the recent work, we apply the PTB-SD representation converted by version 3.3.0 of the Stanford parser.
However, this dependency representation results in around 1\% of phrases containing two or three head words. 
As shown in Figure \ref{Fig.sub.1}, the phrase (5,8) assigned with a category NP contains 2 head words of \textit{paper} and \textit{products} in dependency tree.
In order to deal with the problem, we introduce a special category $\#$ to divide the phrase with multiple heads meeting only one head word for each phrase. After this conversion, only 50 heads are errors in Penn Treebank.

\label{Tree Preprocessing}
\begin{figure*}[t!]
    \centering
    \includegraphics[width=6in]{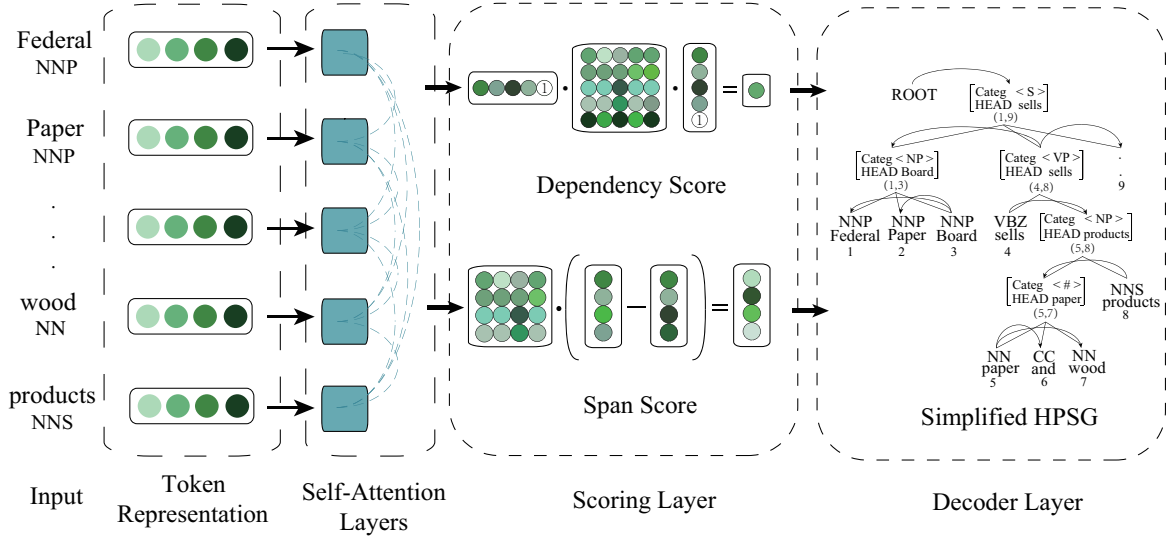}
    
    \caption{The framework of our joint span HPSG parsing model.}
    \label{fig3}
\end{figure*}
\subsection{Span Representations of HPSG}

Each node in the HPSG tree noted as AVM represents compound structure. Even in our simplified HPSG, each phrase (span) should be companied with its head.
To facilitate the processing of existing parsers, we propose two ways to convert the simplified HPSG into a span-style tree structure.

\noindent \textbf{Division Span} A phrase is divided into two parts corresponding to left and right of its head. 
To distinguierrorsh the left and right parts, we add a special token $H$ in front of the category to indicate the left span, in which the head of the original phrase is always the last word.
Since some leaves of the tree are without category, we explicitly use a special empty category $\O$ for their representation, and the token $H$ is also applied to the empty category.

As shown in Figure \ref{Fig.sub.2}, the head of phrase (1,3) in the dotted box is \textit{Board}, thus we add the special token $H$ in front of \textit{Federal},  \textit{Paper} and \textit{Board} category.
With this operation, head information has been encoded into span boundary of a standard constituent tree and we only need to parse such a constituent tree.

\noindent \textbf{Joint Span}
We recursively define a structure called joint span to cover both constituent and head information.
A joint span consists of all its children phrases and all dependency arcs between heads of all these children phrases. 

For example, the HPSG node S$_{H}$(1, 9) in Figure \ref{Fig.sub.3} as a joint span is:
\begin{align*}
S_{H}(1, 9) = \{ S_{H}(1, 3) , S_{H}(4, 8) , S_{H}(9, 9),\\ l(1, 9),
d(Board, sells) , d(., sells) \},
\end{align*}
where $l$($i$, $j$) denotes category of span
($i$, $j$) and $d$($r$, $h$) indicates the dependency
between the word $r$ and its parent $h$. 

At last, following the recursive definition, the entire HPSG tree $T$ being a joint span can be represented as:
\begin{align*}
S_H(T) = \{S_H(1, 9), d(sells, root)\}.
\end{align*}
As all constituent and head information has been formally encoded into a span-like structure, we can use a constituent-like parser for such a joint span tree.

\label{Two structure of Simplified HPSG}

\section{Our Model}

\subsection{Overview}

Using an encoder-decoder backbone, our model apply self-attention encoder \cite{Vaswani17} which is modified by position partition \cite{Kitaev-2018-SelfAttentive}.
Since our two converted structures of simplified HPSG are based on the phrase, thus we can employ CKY-style \cite{Cocke1970Programming, Younger1975Recognition, Kasami1965AN} decoder for both to find the tree with the highest predicted scores.
The difference is that for division span structure, we only need span scores while for joint span structure, we need both of span and dependency scores.

Given a sentence $s = \{w_1, w_2, \dots ,w_n\}$, we attempt to predict a simplified HPSG tree. As shown in Figure \ref{fig3}, our parsing model includes four modules: token representation, self-attention encoder, scoring module and CKY-style decoder\footnote{For dependency label of each word, it is not necessary for our HPSG parsing purpose, however, to enable  our parser fully comparable to existing dependency parsers, we still  train a separated multiclass classifier simultaneously with the parser by optimizing the sum of their objectives.}.

\subsection{Token Representation}

In our model, token representation $x_i$ is composed of character, word and part-of-speech (POS) embeddings.
For character-level representation, we use CharLSTM \cite{Kitaev-2018-SelfAttentive}.
For word-level representation, we concatenate randomly initialized and pre-trained word embeddings.

Finally, we concatenate character representation, word representation
and POS embedding as our token representation:
$$
x_i=[x_{char};x_{word};x_{POS}]. 
$$

\label{Token Representation}

\subsection{Self-Attention Encoder}

The encoder in our model is adapted from \cite{Vaswani17} and factor explicit content and position information in the self-attention process. The input matrices $X = [x_1, x_2, \dots , x_n ]$ in which $x_i$ is concatenated with position embedding are transformed by a self-attention encoder. We factor the model between content and position information both in self-attention sub-layer and feed-forward network, whose setting details follow \cite{Kitaev-2018-SelfAttentive}. 

\subsection{Decoder for Division Span HPSG}

After reconstructing of the HPSG tree as a constituent tree with head information as described in Section \ref{Two structure of Simplified HPSG}, we follow the constituent parsing as \cite{Kitaev-2018-SelfAttentive,Gaddy} to predict constituent parse tree.

Firstly, we add a special empty category $\O$ to spans to binarize the n-ary nodes and apply a unary atomic category to deal with the nodes of the unary chain, corresponding to nested spans with the same endpoints.


Then, we train the span scorer.
Span vector $s_{ij}$ is the concatenation of the vector differences $s_{ij} = [\overrightarrow{y_j}-\overrightarrow{y_{i-1}};\overleftarrow{y_{j+1}}-\overleftarrow{y_{i}}]$ 
which $\overrightarrow{y_j}$ is constructed by splitting in half the outputs from the self-attention encoder. 
We apply one-layer feedforward networks to generate span scores vector, taking span vector $s_{ij}$ as input:
$$
S(i, j) = W_2 g(LN( W_1s_{ij}+ b_1)) + b_2,
$$
where $LN$ denotes Layer Normalization, $g$ is the Rectified Linear Unit nonlinearity.
The individual score of category $\ell$ is denoted by
$$
S_{categ}(i, j, \ell) = [S(i, j)]_\ell, 
$$
where $[]_{\ell}$ indicates the value of corresponding the element $\ell$ of the score vector.
The score $s(T)$ of the constituent parse tree $T$ is to sum every scores of span ($i$, $j$) with category $\ell$: 
$$s(T) = \sum_{(i,j,\ell)\in T} S_{categ}(i, j, \ell).$$
The goal of constituent parsing is to find the tree with the highest score:
$\hat{T} = \arg\max_T s(T).$
We use CKY-style algorithm \cite{SternP17,Gaddy} to obtain the tree $\hat{T}$ in $O(n^3)$ time complexity.
This structured prediction problem is handled with satisfying the margin constraint:
$$
s(T^*) \ge s(T) + \Delta (T,T^*),
$$
 where $T^*$ denotes correct parse tree and $\Delta$ is the Hamming loss on category spans with a slight modification during the dynamic programming search. The objective function is the hinge loss,
$$J_1(\theta) = \max ( 0,\max_T[s(T) + \Delta (T,T^*)]-s(T^*) ).$$

For dependency labels, following \cite{Dozat2017Deep}, the classifier takes head and its children as features.
We minimize the negative log probability of the correct dependency label $l_i$ for the child-parent pair $(x_i,h_i)$ implemented as cross-entropy loss:
$$
J_{labels}(\theta) = -logP_{\theta}(l_i|x_i,h_i).
$$
 Thus, the overall loss is sum of the objectives:
$$
J_{Division}(\theta) = J_1(\theta) + J_{labels}(\theta).
$$

\begin{algorithm}[t!]
  \caption{Joint span parsing algorithm}
  \label{alg1}
  \begin{algorithmic}
  \REQUIRE sentence leng $n$, span and dependency score $s(i,j,\ell)$, $d(r,h)$, $1\leq i\leq j \leq n, \forall r,h,\ell$
  \ENSURE maximum value $S_H(T)$ of tree $T$
  \STATE \textbf{Initialization:} 
  \STATE $s_{c}[i][j][h] = s_{i}[i][j][h] = 0,  \forall i,j,h $ 
  \FOR{$len=1$ to $n$}
    \FOR{$i=1$ to $n-len + 1$}
        \STATE $j = i + len - 1$
        \IF {$len=1$}
            \STATE $\begin{aligned}
            s_{c}[i][j][i] = s_{i}[i][j][i] = \max_{\ell} s(i,j,\ell)
            \end{aligned}$
        \ELSE
        \FOR{$h=i$ to $j$}
            \STATE 
            $\begin{aligned}
                split_l = &\max_{i\leq r < h} \ \{\ \max_{r \leq k < h}\ \{\ s_{c}[i][k][r] + \\
                &s_{i}[k+1][j][h]\ \} + d(r,h)\ \}
            \end{aligned}$
            
            \STATE  $\begin{aligned}
            split_r =& \max_{h < r \leq j}\ \{\ \max_{h \leq k < r}\ \{\ s_{i}[i][k][h] + \\
            &s_{c}[k+1][j][r]\ \} + d(r,h)\ \}
            \end{aligned}$
            
            \STATE $\begin{aligned}
            s_{c}[i][j][h] = & \max \ \{\ split_l, split_r\ \} + \\
            &\max_{\ell \neq \varnothing} s(i,j,\ell)
            \end{aligned}$
            
            \STATE $\begin{aligned}
            s_{i}[i][j][h] = &\max \ \{\ split_l, split_r\ \} + \\
            & \max_{\ell} s(i,j,\ell)
            \end{aligned}$
        \ENDFOR
        \ENDIF
   \ENDFOR
   \ENDFOR
   \STATE $\begin{aligned}
   S_H(T) =  \max_{1 \le h \le n}\ \{\ s_{c}[1][n][h] + d(h,root)\ \}
   \end{aligned}$
  \end{algorithmic}
\end{algorithm}

\subsection{Decoder for Joint Span HPSG}

As our joint span is defined in a recursive way, to score the root joint span has been equally scoring all spans and dependencies in the HPSG tree.

For span scores, we continuously apply the approach and hinge loss $J_1(\theta)$ in the previous section. 
For dependency scores, we predict a distribution over the possible head for each word and use the biaffine attention mechanism \cite{Dozat2017Deep} to calculate the score as follow:
$$
\alpha_{ij} = h_i^TWg_j + U^Th_i + V^T g_j + b,
$$
where $\alpha_{ij}$ indicates the child-parent score, $W$ denotes the weight matrix of the bi-linear term, $U$ and $V$ are the weight vectors of the linear term and $b$ is the bias item, $h_i$ and $g_i$ are calculated by a distinct one-layer perceptron network.

We minimize the negative log-likelihood of the golden dependency tree $Y$, which is implemented as a cross-entropy loss:
$$
J_2(\theta) = -\left(logP_{\theta}(h_i|x_i) +logP_{\theta}(l_i|x_i,h_i)\right),
$$
where $P_{\theta}(h_i|x_i)$ is the probability of correct parent node $h_i$ for $x_i$, and $P_{\theta}(l_i|x_i,h_i)$ is the probability of the correct dependency label $l_i$ for the child-parent pair $(x_i,h_i)$.
To predict span and dependency scores simultaneously, we jointly train our parser for minimizing the overall loss:
$$
J_{Joint}(\theta) = J_1(\theta) + J_2(\theta).
$$
During testing, we propose a CKY-style algorithm as shown in Algorithm \ref{alg1} to explicitly find the globally highest span and dependency score $S_H(T)$ of our simplified HPSG tree $T$. 
In order to binarize the constituent parse tree with head, we 
introduce the complete span $s_c$ and the incomplete span $s_i$ which is similar to Eisner algorithm \cite{EisnerP96}.
After finding the best score $S_H(T)$, we backtrack the chart with split point $k$ and sub-root $r$ to construct the simplified HPSG tree $T$. 

Comparing with constituent parsing CKY-style algorithm \cite{SternP17}, the dependency score $d(r,h)$ in our algorithm affects the selection of best split point $k$.
Since we need to find the best value of sub-head $r$ and split point $k$,
the complexity of the algorithm is $O(n^5)$ time and $O(n^3)$ space.
To control the effect of combining span and dependency scores, we apply a weight $\lambda$:
$$
s(i,j,\ell) = \lambda  S_{categ}(i,j,\ell),
d(i,j) = (1.0 - \lambda )  \alpha_{ij},
$$
where $\lambda$ in the range of 0 to 1.
In addition, we can merely generate constituent or dependency parsing tree by setting $\lambda$ to 1 or 0, respectively.
\label{Joint Span Decoder}

\section{Experiments}

In order to evaluate the proposed model, we convert our simplified HPSG tree to constituent and dependency parse trees and evaluate on two benchmark treebanks, English Penn Treebank (PTB) and Chinese Penn Treebank (CTB5.1) following standard data splitting \cite{ZhangD08, Liuandzhang2017B}. The placeholders with the -NONE- tag are stripped from the CTB. 
POS tags are predicted using the Stanford tagger \cite{Toutanova:2003} and we use the same pretagged dataset as \cite{Cross} for PTB.
For CTB, we use golden POS tags for dependency parsing and predicted POS tags for constituent parsing.

For constituent parsing, we use the standard evalb{\footnote{http://nlp.cs.nyu.edu/evalb/}} tool to evaluate the F1 score. For dependency parsing, following \cite{Dozat2017Deep, Kuncoro2016Distilling, Ma2018Stack}, we report the results without punctuations for both treebanks.

\subsection{Setup}

\textbf{Hyperparameters} In our experiments, we use 100D GloVe \cite{PenningtonD14-1162} and structured-skipgram \cite{LingN15-1142} pre-train embeddings for English and Chinese respectively. 
The character representations are randomly initialized, and the dimension is 64. For self-attention encoder, we use the same hyperparameters settings as \cite{Kitaev-2018-SelfAttentive}.

For span scores, we apply a hidden size of 250-dimensional feed-forward networks.
For dependency biaffine scores, we employ two 1024-dimensional MLP layers with the ReLU as the activation function and a 1024-dimensional parameter matrix for biaffine attention. 
In addition, we augment our parser with ELMo \cite{PetersN18-1202}, a larger version of BERT \cite{Jacobbert} (24 layers, 16 attention heads per layer, and 1024-dimensional hidden vectors) and XLNet \cite{XLNet-Zhilin-2019} to compare with other pre-trained or ensemble models.
We set 4 layers of self-attention for ELMo and 2 layers of self-attention for BERT or XLNet as \cite{Kitaev-2018-SelfAttentive, kitaev2018multilingual}.

\noindent \textbf{Training Details} we use 0.33 dropout for biaffine attention and MLP layers. All models are trained for up to 150 epochs with batch size 150 on a single NVIDIA GeForce GTX 1080Ti GPU with Intel i7-7800X CPU. 
We use the same training settings as \cite{Kitaev-2018-SelfAttentive} and \cite{kitaev2018multilingual}.

\begin{table}[t!]
    \centering
    \resizebox{\linewidth}{!}{  
        \begin{tabular}{lccl}
            \hline
            \bf Self-attention Layers      &F1 &UAS &LAS\\
            \hline
            Division Span Model\\
            \hline
            8 self-attention layers      &93.42 &94.05 &92.68 \\
            12 self-attention layers &\bf93.57 &\bf94.40 &\bf93.05 \\
            16 self-attention layers   &93.36 &94.08 &92.66 \\
            \hline
            \hline
            Joint Span Model\\
            \hline
            8 self-attention layers      &93.64 &95.75 &94.36 \\
            12 self-attention layers   &\bf93.78 &\bf95.92 &\bf94.49 \\
            16 self-attention layers   &93.54 &95.54 &94.21 \\
            \hline
        \end{tabular}}
    \caption{\label{table1} Different self-attention layers on English dev set.}
\end{table}

\subsection{Self-attention Layers} 

This subsection examines the impact of different numbers of self-attention layers varying from 8 to 16.
The comparison in Table \ref{table1} indicates that the best performing setting comes from 12 self-attention layers, and more than 12 layers shows almost no promotion even reduces the accuracy.
Thus the rest experiments are done with 12 layers of the self-attention encoder.

\subsection{Moderating constituent and Dependency}

The weight parameter $\lambda$ plays an important role to balance the scoring of span and dependency.
When $\lambda$ set to 0, indicates only using dependency score to generate dependency tree as the general first-order dependency parsing \cite{EisnerP96}, while $\lambda$ set to 1, shows the constituent parsing only.
$\lambda$ set to between 0 to 1 indicates our general simplified HPSG parsing, providing both constituent and dependency structure prediction.

\begin{figure}[t!]
    \centering
    \includegraphics[width=3in]{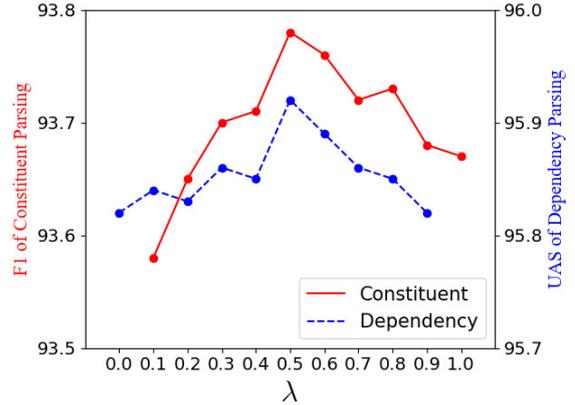}
    
    \caption{Balancing constituent and dependency of joint span HPSG parsing on English dev set.}
    \label{fig4}
\end{figure}

The comparison in Figure \ref{fig4} shows that our HPSG decoder is better than either separate constituent or dependency decoder, which shows the bonus of joint predicting constituent and dependency.
Moreover, $\lambda$ set to 0.5 achieves the best performance in terms of both F1 score and UAS.

\begin{table}[t!]
    \centering
    \resizebox{\linewidth}{!}{  
    
        \begin{tabular}{lccc}
            \hline
            \bf Model                           &F1 &UAS &LAS\\
            \hline
            separate constituent      &\multirow{2}{0.7cm}{93.47} &- &- \\
            converted dependency      &         &95.06 &93.81 \\
            \hline
            joint span $\lambda$ = 1.0   &93.67 &-  &- \\
            joint span $\lambda$ = 0.0   &-  &95.82 &94.43 \\
            joint span $\lambda$ = 0.5   &\multirow{2}{0.7cm}{\bf93.78} &\bf95.92 &\bf94.49 \\
            converted dependency   &  &95.69 &94.45 \\
            \hline
        \end{tabular}}
    \caption{\label{table3} English dev set performance of joint span HPSG parsing. The \textit{converted} means the corresponding dependency parsing results are from the corresponding constituent parse tree using head rules.}
\end{table}

\subsection{Joint Span HPSG Parsing}

We compare our join span HPSG parser with a separate learning constituent parsing model which takes the same token representation and self-attention encoder on PTB dev set.
The constituent parsing results are also converted into dependency ones by PTB-SD for comparison.

When $\lambda$ is set to 0 and 1, our joint span HPSG parser works as the dependency-only parser and constituent-only parser respectively. Table \ref{table3} shows that even in such a work mode, our HPSG parser still outperforms the separate constituent parser in terms of either constituent and dependency parsing performance.

As $\lambda$ is set to 0.5, our HPSG parser will give constituent and dependency structures at the same time, which are shown better than the work alone mode of either constituent or dependency parsing. Besides, the comparison also shows that the directly predicted dependencies from our model are slightly better than those converted from the predicted constituent parse trees.
\begin{table}[t!]
    \centering
    \resizebox{\linewidth}{!}{  
    \begin{tabular}{lr}
            \hline
            \bf Model      & sents/sec\\
            \hline
            \citet{PetrovN07-1051}      &6.2 \\
            \citet{ZhuZhang2013}        &89.5 \\
            \citet{Liuandzhang2017B}  &79.2 \\
            \citet{SternP17}            &75.5 \\
            \citet{ShenP18}             &111.1 \\
            \citet{ShenP18}(w/o tree inference) &351 \\
            \hline
            \hline
            Our (Division)    &226.3 \\
            Our (Joint)       &158.7   \\
            \hline
        \end{tabular}}
    \caption{\label{table3} Parsing speed on the PTB dataset.}
\end{table}

\begin{table}[t!]
    \centering
    \resizebox{\linewidth}{!}{  
        \begin{tabular}{l|ll|ll}
            \hline
            \multirow{2}{*}{\bf Model} & \multicolumn{2}{c|}{English} & \multicolumn{2}{c}{Chinese} \\
            \cline{2-5}
            \multirow{2}{*}&UAS &LAS &UAS &LAS\\
            \hline
            \citet{ChenD14} &91.8 &89.6 &83.9 &82.4\\
            \citet{Andor2016Globally} &94.61 &92.79 &\_ &\_\\
            \citet{zhang-etal-2016-probabilistic} &93.42 &91.29 &87.65 &86.17\\
            \citet{ChengD16-1238} &94.10 &91.49 &88.1 &85.7\\
            \citet{Kuncoro2016Distilling} &94.26 &92.06 &88.87 &87.30\\
            \citet{MaI17-1007} &94.88 &92.98 &89.05 &87.74\\
            \citet{Dozat2017Deep} &95.74 &94.08 &89.30 &88.23\\
            \citet{li-etal-2018-seq2seq} &94.11 &92.08 &88.78 &86.23\\
            \citet{Ma2018Stack}  &95.87 &94.19 &90.59 &\bf89.29\\
            Pointer Networks &96.04 &94.43 &- &- \\
            \hline
            Our (Division) &94.32 &93.09 &89.14 &87.31 \\
            Our (Joint)  &\bf96.09 &\bf94.68 &\bf91.21 &89.15 \\
            \hline
            Our (Division*) &- &- &91.69 &90.54 \\
            Our (Joint*)  &- &- &93.24 &91.95 \\
            \hline
            \hline
            \bf Pre-training/Ensemble \\
            \hline
            \citet{ChoeD16-1257}  &95.9  &94.1 &\_ &\_      \\
            \citet{KuncoroE17-1117} &95.8  &94.6 &\_ &\_    \\
            \citet{WangD18-1311}(ELMo) &96.35 &95.25 &\_ &\_ \\
            
            \hline
            Our (Division) + ELMo &95.77 &94.21 &- &- \\
            Our (Joint) + ELMo  &96.76 &94.93 &- &- \\
            Our (Division) + BERT &96.22 &94.56 &- &- \\
            Our (Joint) + BERT  &97.00 &95.43 &- &- \\
            Our (Joint) + XLNet  &\bf97.20 &\bf95.72 &- &- \\
            \hline
        \end{tabular}}
    \caption{\label{table4} Dependency parsing on PTB and CTB test set. * represents CTB constituent data splitting. Pointer Networks represents the results of \citet{Daniel-2019-naacl-left}.}
\end{table}

\subsection{Parsing Speed}

We compare the
parsing speed of our parser with other neural parsers in Table \ref{table3}.
Although the time complexity of our \textit{Joint} span model is $O(n^5)$, there is not much slower than \textit{Division} span model with $O(n^3)$ time complexity.
The comparison suggests that training and inference
times are dominated by neural network computations and our decoder consumes a small fraction of total running time.

\subsection{Main Results}

Tables \ref{table4}, \ref{table5} and \ref{table6} compare our model to existing state-of-the-art on test sets.
\textit{Division} and \textit{Joint} indicate the results of division and joint span parsing respectively.
On PTB, our best model achieves new state-of-the-art on both constituent and dependency parsing.
On CTB, our best model achieves 89.40 F1 score of constituent parsing and 91.21\% UAS and 89.15\% LAS of dependency parsing.
Since constituent and dependency parsing have different data splitting on CTB \cite{ZhangD08, Liuandzhang2017B}, we report our parsing performance on both data splitting.

The comparison shows that our HPSG parsing model is more effective than learning constituent or dependency parsing separately.
We also find that dependency parsing is shown much more beneficial from \textit{Joint} than \textit{Division} way which empirically suggests dependency score in our joint loss is helpful.

We augment our parser with ELMo, a larger version of BERT and XLNet as the sole token representation to compare with other models.
Our \textit{Joint} model in XLNet setting even defeats other ensemble models of both constituent and dependency parsing achieving 96.33 F1 score, 97.20\% UAS and 95.72\% LAS.

For fair comparison with other pre-train model on constituent parsing, we also augment our parser with Chinese larger version of RoBERTa\footnote{https://github.com/brightmart/roberta\_zh} as the sole token representation.
Our \textit{Joint} model in RoBERTa setting achieves the state of art performance of 92.55 F1 score on constituent parsing.

\begin{table}[t!]
    \centering
        \resizebox{\linewidth}{!}{
        \begin{tabular}{llll}
            \hline
            \bf Model               &LR &LP &F1\\
            \hline 
            \citet{ZhuZhang2013} &90.7 &90.2 &90.4\\
            \citet{Dyer-N16-1024} &\_ &\_ &89.8\\
            \citet{Cross} &90.5 &92.1 &91.3\\
            \citet{SternP17}&93.2 &90.3 &91.8\\
            \citet{Gaddy} &91.76 &92.41 &92.08\\
            \citet{SternD17b} &92.57 &92.56 &92.56\\
            \citet{Kitaev-2018-SelfAttentive}  &93.20  &93.90 &93.55\\
            \hline
            Our (Division)     &93.41  &93.87 &93.64 \\
            Our (Joint)    &\bf93.64  &\bf93.92 &\bf93.78 \\
            \hline
            \hline
            \bf Pre-training/Ensemble \\
            \hline
            \citet{Dyer-N16-1024} &\_ &\_ &93.3\\
            \citet{ChoeD16-1257} &\_ &\_ &93.8\\
            \citet{LiuandZhang2017A} &\_ &\_ &94.2\\
            \citet{Fried2017Improving} &\_ &\_ &94.66\\
            \citet{Kitaev-2018-SelfAttentive} & & & \\
            + ELMo &94.85 &95.40 &95.13  \\
            \citet{kitaev2018multilingual} & & & \\
            + BERT &95.46 &95.73 &95.59 \\
            \citet{kitaev2018multilingual} &95.51 &96.03 &95.77\\
            \hline
            Our (Division) + ELMo     &94.54  &95.68 &95.10 \\
            Our (Joint) + ELMo    &95.04  &95.39 &95.22 \\
            Our (Division) + BERT     &95.51  &95.93 &95.72 \\
            Our (Joint) + BERT   &95.70  &95.98 &95.84 \\
            Our (Joint) + XLNet   &\bf96.21  &\bf96.46 &\bf96.33 \\
            \hline
        \end{tabular}
        }
    \caption{\label{table5} Constituent parsing on PTB test set.}
\end{table}

\begin{table}[t!]
    \centering
        \resizebox{\linewidth}{!}{
    \begin{tabular}{llll}
        \hline
        \bf Model              &LR &LP &F1\\
        \hline 
        \citet{Wang2015}    &\_ &\_ &83.2\\
        \citet{Dyer-N16-1024}&\_ &\_ &84.6\\
        \citet{Liuandzhang2017B} &85.9 &85.2 &85.5\\
        \citet{LiuandZhang2017A} &\_ &\_ &86.1\\
        \citet{ShenP18}         &86.6 &86.4 &86.5 \\
        \citet{FriedP18}            &\_ &\_ &87.0 \\
        \citet{TengC18-1011}    &87.1 &87.5 &87.3\\
        \citet{kitaev2018multilingual} &91.55 &91.96 &91.75\\
        \hline 
        Our (Division)     &88.75  &89.79 &89.27 \\
        Our (Joint)    &89.09  &89.70 &89.40 \\
        + RoBERTa &\bf92.50 &\bf92.61 &\bf92.55 \\
        \hline 
        Our (Division*)     &87.41  &88.07 &87.74 \\
        Our (Joint*)    &87.87  &87.96 &87.92 \\
        + RoBERTa &90.14 &89.89 &90.02 \\ 
        \hline
        \end{tabular}}
    \caption{\label{table6} Constituent parsing on CTB test set. * represents CTB dependency data splitting.}
\end{table}

\section{Related Work}

In the earlier time, linguists and NLP researchers discussed how to encode lexical dependencies in phrase structures, like lexicalized tree adjoining grammar (LTAG) \cite{SCHABESC88-2121}
and head-driven phrase structure grammar (HPSG) \cite{pollard1994head} which is a constraint-based highly lexicalized non-derivational generative grammar framework. 

In the past decade, 
there was a lot of large-scale HPSG-based NLP parsing systems which had been built. Such as the \textit{Enju} English and Chinese parser \cite{Miyao2004CorpusOrientedGD, YuC10-2162}, the \textit{Alpino} parser for Dutch \cite{van2006last}, and the \textit{LKB \& PET} \cite{copestake2002implementing, Callmeier2000PET} for English, German, and Japanese..

Meanwhile, since HPSG represents the grammar framework in a precisely constrained way, it is difficult to broadly cover unseen real-world texts for parsing. 
Consequently, according to \cite{Zhang-W11-2923}, 
many of these large-scale grammar implementations are forced to choose to either compromise the linguistic preciseness or to accept the low coverage in parsing. 
Previous works of HPSG approximation focus on two major approaches: grammar based approach \cite{kiefer2004context}, and the corpus-driven approach \cite{krieger2007ubgs} and \cite{Zhang-W11-2923} which proposes PCFG approximation as a way to alleviate some of these issues in HPSG processing.

Recently, with the impressive success of deep neural networks in a wide range of NLP tasks \cite{li-etal-2018-unified, zhang-etal-2018-exploring, li-etal-2018-joint-learning, zhang-etal-2018-modeling, zhang-etal-2018-subword, zhang-zhao-2018-one, cai-etal-2018-full, he-etal-2018-syntax, xiao2019latt, AAAI1816060, 8360031, wang-etal-2018-dynamic, wang-etal-2017-instance, wang-etal-2017-sentence}, constituent and dependency parsing have been well developed with neural network. These models attain state-of-the-art results for dependency parsing \cite{ChenD14, Dozat2017Deep, Ma2018Stack} and constituent parsing \cite{Dyer-N16-1024, Cross, Kitaev-2018-SelfAttentive}. 

Since constituent and dependency share a lot of grammar and machine learning characteristics, it is a natural idea to study the relationship between constituent and dependency structures, and the joint learning of constituent and dependency parsing \cite{CollinsP97-1003,CharniakA00-2018,CharniakP05-1022,FarkasW11-2924,GreenW12-0503,RenCombine2013,YoshikawaP17-1026}.

To further exploit both strengths of the two representation forms, in this work, for the first time, we propose a graph-based parsing model that formulates constituent and dependency structures as simplified HPSG.

\section{Conclusions}

This paper presents a simplified HPSG with two different decode methods which are evaluated on both constituent and dependency parsing. Despite the usefulness of HPSG in practice and its theoretical linguistic background, our model achieves new state-of-the-art results on both Chinese and English benchmark treebanks of both parsing tasks. Thus, this work is more than proposing a high-performance parsing model by exploring the relation between constituent and dependency structures. 
Our experiments show that joint learning of constituent and dependency is indeed superior to separate learning mode, and combining constituent and dependency score in joint training to parse a simplified HPSG can obtain further performance improvement.

\bibliography{acl2019}
\bibliographystyle{acl_natbib}

\end{document}